%% file: camera-ready.tex
\documentclass[10pt,twocolumn,letterpaper]{article}

\usepackage{cvpr}              

\usepackage{graphicx}
\usepackage{amsmath}
\usepackage{amssymb}
\usepackage{booktabs}

\usepackage[misc]{ifsym}
\usepackage[pagebackref,breaklinks,colorlinks]{hyperref}
\usepackage{threeparttable}
\usepackage{multirow}
\usepackage{bbding}
\usepackage[table]{xcolor}
\usepackage{amsthm,amsmath,amssymb}
\usepackage{mathrsfs}
\usepackage{arydshln}
\usepackage{times}
\usepackage{epsfig}
\usepackage{graphicx}
\usepackage{animate}
\usepackage{amsmath}
\usepackage{amssymb}
\usepackage{pifont}
\usepackage{booktabs}
\usepackage{pifont}

\usepackage[capitalize]{cleveref}
\crefname{section}{Sec.}{Secs.}
\Crefname{section}{Section}{Sections}
\Crefname{table}{Table}{Tables}
\crefname{table}{Tab.}{Tabs.}

\def\cvprPaperID{1582} 
\def\confName{CVPR}
\def\confYear{2022}

\begin{document}

\title{\vspace{-20pt}
\resizebox{\textwidth}{!}{TimeReplayer: Unlocking the Potential of Event Cameras for Video Interpolation}}

\author{Weihua He$^{1}$ \qquad Kaichao You$^{2,4}$\qquad Zhendong Qiao$^{1}$\qquad Xu Jia$^{3}$\Letter\qquad Ziyang Zhang$^{2}$\Letter \\Wenhui Wang$^{1}$\Letter\qquad Huchuan Lu$^{3,5}$\qquad Yaoyuan Wang$^{2}$\qquad Jianxing Liao$^{2}$\\
$^{1}$Department of Precision Instrument, Tsinghua University \\
$^{2}$Advanced Computing and Storage Lab, Huawei Technologies Co. Ltd\\
$^{3}$School of Artificial Intelligence, Dalian University of Technology\\
$^{4}$School of Software, Tsinghua University\quad $^{5}$Peng Cheng Laboratory\\
\tt\small \{hwh20@mails,wwh@mail\}@.tsinghua.edu.cn,youkaichao@gmail.com\\
\tt\small \{xjia,lhchuan\}@dlut.edu.cn,\{zhangziyang11,wangyaoyuan1,liaojianxing\}@huawei.com
}

\maketitle

\begin{abstract}
   Recording fast motion in a high FPS (frame-per-second) requires expensive high-speed cameras. As an alternative, interpolating low-FPS videos from commodity cameras has attracted significant attention. If only low-FPS videos are available, motion assumptions (linear or quadratic) are necessary to infer intermediate frames, which fail to model complex motions. Event camera, a new camera with pixels producing events of brightness change at the temporal resolution of $\mu s$ $(10^{-6}$ second $)$, is a game-changing device to enable video interpolation at the presence of arbitrarily complex motion. Since event camera is a novel sensor, its potential has not been fulfilled due to the lack of processing algorithms. The pioneering work Time Lens introduced event cameras to video interpolation by designing optical devices to collect a large amount of paired training data of high-speed frames and events, which is too costly to scale. To fully unlock the potential of event cameras, this paper proposes a novel TimeReplayer algorithm to interpolate videos captured by commodity cameras with events. It is trained in an unsupervised cycle-consistent style, canceling the necessity of high-speed training data and bringing the additional ability of video extrapolation. Its state-of-the-art results and demo videos in supplementary reveal the promising future of event-based vision.
\end{abstract}
\vspace{-0.5cm}

\section{Introduction}
\footnotetext{Corresponding authors: Xu Jia, Ziyang Zhang, and Wenhui Wang.}
\footnotetext{\scriptsize Webpage: \url{https://sites.google.com/view/timereplayer}.}
Traditionally, recording fast motion in high temporal resolution has been the exclusive functionality of high-speed cameras, which are too expensive to be embedded in personal devices such as smartphones. Video interpolation, the research field of inferring intermediate frames between two given frames, has attracted considerable research interest. It can go beyond the temporal resolution limit of commodity cameras, and can be applicable in diverse downstream tasks such as slow motion generation \cite{jiang2018super,xu2019quadratic,bao2019depth}, video editing \cite{zitnick2004high,ren2018deep}, and virtual reality \cite{anderson2016jump}.
\begin{figure}[t]
     \centering
\animategraphics[width=\linewidth, autoplay=True]{4}{figs/video/name_}{0}{11}
\includegraphics[width=0.48\textwidth]{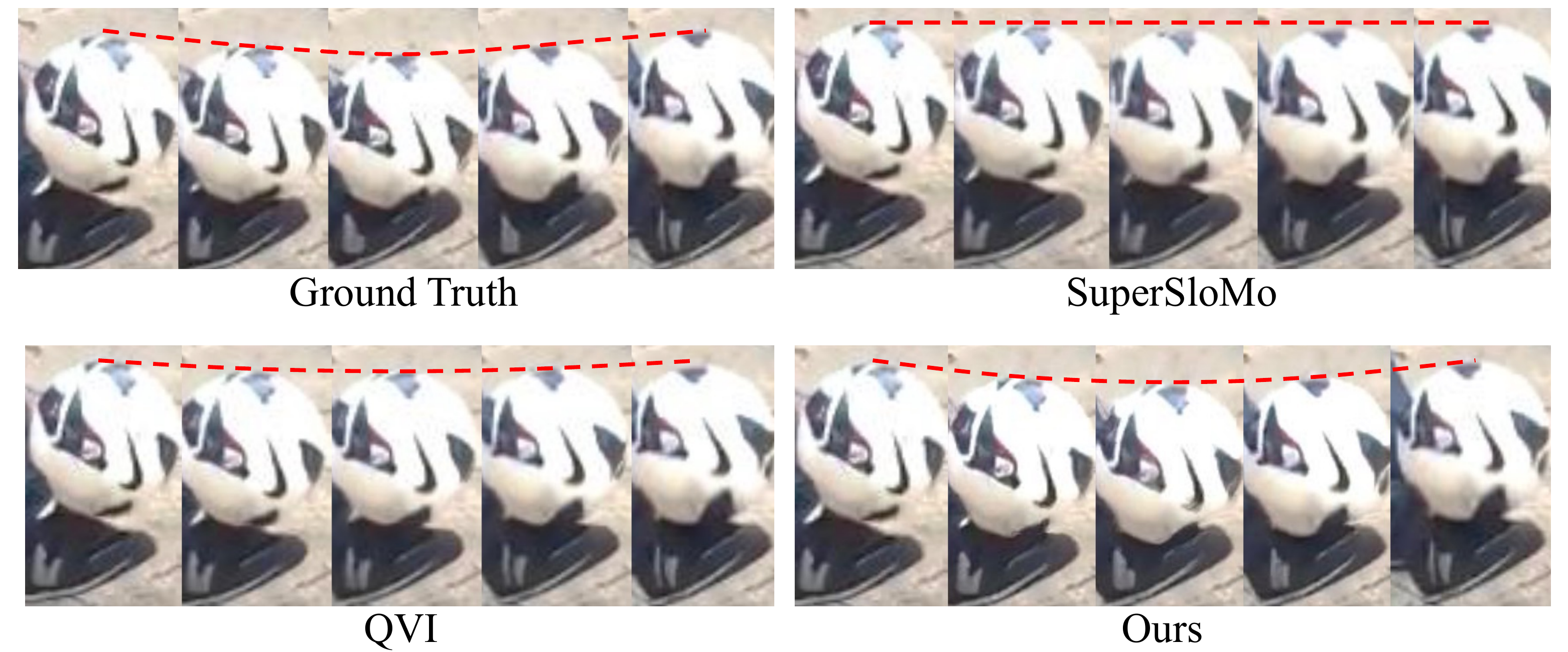}
\caption{Visual results of interpolating complex motion. Both models with linear motion assumption (SuperSloMo\cite{jiang2018super}) and quadratic motion assumption (QVI\cite{xu2019quadratic}) fail to model the motion of the soccer ball correctly. Benefiting from the high temporal resolution of event data, the proposed method can correctly model nonlinear motion and interpolate intermediate frames. \textbf{This figure contains animations. Best viewed in Acrobat Reader.}
\label{fig:soccer_animation}
}

\vspace{-0.5cm}
\end{figure}
Without any further information or proper assumption, the video interpolation problem is under-determined. In the past years, the computer vision community has explored a lot in the direction of exploiting proper assumptions. \cite{jiang2018super} proposes SuperSloMo for video interpolation with linear motion assumption, and \cite{xu2019quadratic} improves the result based on a quadratic motion assumption. As the assumption becomes more and more complicated, two obstacles get in the way: (1) linear motion~\cite{jiang2018super} can interpolate video given two frames, but quadratic motion~\cite{xu2019quadratic} requires four consecutive frames to calculate the acceleration for quadratic interpolation. As the assumption becomes more complex, more frames and more computational costs are needed to interpolate every frame. (2) the underlying motion in the video can be arbitrary, making it difficult to verify pre-set assumptions on the motion type. When the assumption and the motion type mismatch, the interpolated video may look unrealistic. 

To deal with the inherent lack of intermediate information of conventional frame-based cameras, we bring in a novel neuromorphic sensor, event camera\cite{delbruck2008frame,lichtsteiner2008128,serrano2013128,yang2015dynamic}, as a promising solution for video interpolation in case of complex motion. 
The event camera senses the dynamic change of pixel intensity, where an event is triggered once the change exceeds a certain threshold. The asynchronously event-driven processing fashion leads to a frame-less event stream with microsecond temporal resolution, as well as low power consumption and bandwidth \cite{pei2019towards,he2020comparing}. 
Since event camera almost records continuously streaming intensity change, it is able to store rich inter-frame information, which is critical to recovering intermediate video frames. The introduction of event camera for video interpolation helps ease the difficulty in modeling complex motion. 


\begin{table*}[htbp]
   \centering
   \caption{Overview of video interpolation methods, including the proposed TimeReplayer in this paper.}
   \vspace{-5pt}
   \resizebox{.8\textwidth}{!}{
     \begin{tabular}{clllc}
      \toprule
      Input Modality
      & Method & Motion assumption & Training data collection & Supervision \\
           \midrule
            \multirow{3}[0]{*}{RGB frames} & SuperSloMo~\cite{jiang2018super} & linear motion assumption & high speed camera & supervised \\
            & QVI~\cite{xu2019quadratic}   & quadratic motion assumption & high speed camera & supervised \\
            & UnSuperSloMo~\cite{reda2019unsupervised}   & linear motion assumption & high speed camera & unsupervised \\
           \midrule
            RGB frames & TimeLens~\cite{tulyakov_time_2021} & no motion assumption & high speed camera + event camera & supervised \\
            + Events & TimeReplayer & no motion assumption & commodity camera+ event camera & unsupervised \\
           \bottomrule
     \end{tabular}
   }
   \label{tab:addlabel}%
 \end{table*}%
 
Another crucial issue with the task of video interpolation is that it is difficult to collect ground truth for video interpolation in the real world scenarios. Most existing video interpolation models~\cite{jiang2018super, bao2019depth, xu2019quadratic,niklaus2017video} are trained on high-frame-rate dataset, which is constructed by taking an average of consecutive frames of high-frame-rate videos recorded by high-speed cameras. However, these models would suffer from the domain gap between synthetic data and real world data. Therefore, it is important to equip the models with the learning ability to generalize to data without ground truth. 
There have been a few works~\cite{liu2019deep, reda2019unsupervised} aiming to address this issue. They base their methods on cycle consistency, where multiple intermediate frames are predicted and are then used to reconstruct a middle input frame. 
However, to achieve cycle consistency, these methods have to assume uniform motion between consecutive frames at large timesteps, which makes them also suffer from the same problems as those uniform motion assumption-based methods. 

In this work we introduce data from event camera into the video interpolation model design and propose an unsupervised learning framework to train the video interpolation model.
Specifically, we introduce event stream to help directly estimate optical flows between an intermediate frame and input frames instead of computing them as proportions of the computed optical flow between input frames. In this way the uniform linear motion assumption is broken and the proposed optical flow estimation module can compute any complex nonlinear motion. Then an intermediate frame can be predicted by warping input frames according to the estimated optical flows and taking a weighted average of them. 
With the help of event streams and approximated inverse ones, a video interpolation model is able to predict an intermediate frame given two input frames and can also reconstruct an input frame when the estimated intermediate frame and another input frame are given. Therefore, loss functions can be computed between the reconstruction of input frames and the original ones and are used to train the video interpolation model. Extensive experiments and ablation study on synthetic benchmark datasets demonstrate the effectiveness of the proposed framework, especially on videos with complex motion. In addition, we also further conduct an experiment on real data to validate its generalization ability.
%
Contributions of our work can be summarized as below.
\begin{itemize}
   \item A novel video frame interpolation method with both video frame and event stream as input is proposed to address complex nonlinear motion.
   \item We design an unsupervised learning framework for video interpolation with event stream by applying cycle consistency. 
   \item The proposed method performs favorably against state-of-the-art approaches on both synthetic  benchmarks datasets and real data. When trained with additional unsupervised data, it achieves the best result thanks to the unsupervised nature, showing the promising future of event-based video interpolation.
\end{itemize}

\section{Related Work}

\textbf{Video Frame interpolation} Current state-of-the-art interpolation methods are mainly based on warping bidirectional optical flow. However, most of them \cite{bao2019memc,jiang2018super,liu2017video,meyer2015phase,reda2019unsupervised} simply assumed uniform motion and linear optical flow between consecutive frames, which may fail in approximating the complex and nonlinear movement in the real world. Thus, the nonlinear QVI methods \cite{xu2019quadratic,liu2020enhanced} are proposed to solve this problem by learning higher-order acceleration information between frames. But the errors in acceleration estimation may lead to deviation from the motion trajectory in the ground truth. Another essential problem in video interpolation is there are no paired ground truth frames for model supervision in the low-frame-rate video. Therefore, fine-tuning of the supervised method is impossible to be performed on target low-frame-rate image sequence, and the model may suffer performance degradation when applied to the target source. Unsupervised interpolation method \cite{reda2019unsupervised} is established to overcome the problem, obtaining supervision from original input frames based on cycle consistency. However, limitations exist since the method assumes longer temporal consistency in three consecutive frames, further exacerbating the disadvantages of the linear models when facing complexity in real world.

\textbf{Event Camera} 
As a neuromorphic sensor, event camera \cite{delbruck2008frame,lichtsteiner2008128,serrano2013128,yang2015dynamic} asynchronously triggers events once the log intensity change of per-pixel reaches threshold. The unique working pipeline brings high temporal resolution (1us), high dynamic range (140dB), and low power consumption. Previous research has established that the event-driven can achieve great advantages in the image and video synthesis, thus various event-driven algorithms have been developed for downstream visual tasks such as motion deblurring \cite{jiang2020learning,lin2020learning,pan2019bringing}, optical flow estimation \cite{zhu2018ev,lee2020spike}, high dynamic range imaging \cite{han2020neuromorphic} and visual tracking \cite{yang2019dashnet}. One of the branches \cite{rebecq2019high,munda2018real} is to reconstruct event streams as intensity images to recover high-frame-rate image sequence. However, the corrupted event signals affect the quality of the reconstructed images. With the development of a novel event camera, DAVIS \cite{brandli2014240}, the event channel and the conventional intensity channel are integrated and share the same photo-sensor array, thus avoiding the registration problem and enabling complementarity between event cameras and conventional intensity cameras. By simultaneously using intensity image and high-speed event stream, we can achieve better visual performance in interpolation tasks. Another high-frame-rate synthesis branch \cite{jiang2020learning,lin2020learning,pan2019bringing} is to recover multiple sharp images from the motion-blurring frames. However, the sequence can be reconstructed only within the temporal range of obtained image information by the intensity channel, which is powerless for moments that are not covered by intensity exposure.

\textbf{Cycle Consistency}
Cycle consistency has been widely applied to establish constraints in absence of the direct supervision, such as 3d dense correspondences \cite{zhou2016learning}, disambiguating visual relations \cite{zach2010disambiguating}, and unpaired image-to-image translation \cite{zhu2017unpaired}. When challenged with interpolation task, the fully unsupervised method \cite{reda2019unsupervised} based on cycle consistency could learn behaviors and synthesize high-frame-rate interpolation from any target lower-frame-rate video sequence. As the most related method to our work, the previous unsupervised method \cite{reda2019unsupervised} generates two intermediate frames between three consecutive frames, and then interpolate a frame using the generated frames to match with the intermediate original input frame. 
However, to achieve cycle consistency, it has to use multiple input frames and  assume uniform motion between consecutive frames at large timesteps, which makes them suffer from artifacts caused by inaccurate motion prediction. 

\input{pre_and_method_v2}
\begin{table*}[!t]
   \renewcommand\arraystretch{1.3}
   \centering
   \caption{Quantitative comparison on Adobe240, GoPro, Middlebury (other) and Vimeo90k (interpolation) dataset with synthetic events.}
   \vspace{-5pt}
   \resizebox{\textwidth}{!}{
   \begin{tabular}{lcccccc|ccc|ccc|ccc}
\toprule
                         &            &              &                         &\multicolumn{6}{c|}{Adobe240}                                             & \multicolumn{6}{c}{GoPro}   \\ \cline{5-16} 
\multirow{-1}{*}{Method} & \multirow{-1}{*}{Frame} & \multirow{-1}{*}{Event}& \multirow{-1}{*}{Supervision}& \multicolumn{3}{c|}{7 skip (whole)} & \multicolumn{3}{c|}{7 skip (center)} & \multicolumn{3}{c|}{7 skip (whole)} & \multicolumn{3}{c}{7 skip (center)}                                                \\ \cline{5-16} 
                         &            &             &                         & PSNR$\uparrow$       & SSIM$\uparrow$      & IE$\downarrow$        & PSNR$\uparrow$       & SSIM$\uparrow$       & IE$\downarrow$        & PSNR$\uparrow$       & SSIM$\uparrow$      & IE$\downarrow$        & PSNR$\uparrow$       & SSIM$\uparrow$       & IE$\downarrow$       \\ \hline
E2VID\cite{rebecq2019high}                    &         \ding{56}                &   \ding{52}&   \ding{52}                      & 10.40      & 0.570     & 75.21     & 10.32      & 0.573      & 76.01     & 9.74       & 0.549     & 79.49     & 9.88       & 0.569      & 80.08     \\
SepConv\cite{niklaus2017video}                  &        \ding{52}                 &      \ding{56} &   \ding{52}                   & 32.31      & 0.930     & 7.59      & 31.07      & 0.912      & 8.78      & 29.81      & 0.913     & 8.87      & 28.12      & 0.887      & 10.78     \\
QVI \cite{xu2019quadratic}                     &            \ding{52}             &         \ding{56} &   \ding{52}               & 32.87      & 0.939     & 6.93      & 31.89      & 0.925      & 7.57      & 31.39      & 0.931     & 7.09      & 29.84      & 0.911      & 8.57      \\
DAIN\cite{bao2019depth}                     &             \ding{52}            &         \ding{56}&   \ding{52}                & 32.08      & 0.928     & 7.51      & 30.31      & 0.908      & 8.94      & 30.92      & 0.901     & 8.60      & 28.82      & 0.863      & 10.71     \\
SuperSloMo\cite{jiang2018super}               &            \ding{52}             &          \ding{56} &   \ding{52}              & 31.05      & 0.921     & 8.19      & 29.49      & 0.900      & 9.68      & 29.54      & 0.880     & 9.36      & 27.63      & 0.840      & 11.47     \\
Time Lens\cite{tulyakov_time_2021}                &             \ding{52}            &      \ding{52}  &   \ding{52}                 & 35.47      & 0.954     & 5.92      & 34.83      & 0.949      & 6.53      & 34.81      & 0.959     & 5.19      & 34.45      & 0.951      & 5.42      \\
\hdashline[4pt/5pt]
UnSuperSloMo\cite{reda2019unsupervised}             &            \ding{52}             &      \ding{56}  &   \ding{56}                 & 29.92      & 0.908     & 9.10      & 29.36      & 0.898      & 9.85      & 28.23      & 0.861     & 10.35     & 27.32      & 0.836      & 11.67     \\
Ours                     &           \ding{52}              &       \ding{52} &   \ding{56}                 & 34.14      & 0.950     & 6.25      & 33.22      & 0.942      & 6.64      & 34.02      & 0.960     & 5.02      & 33.39      & 0.952      & 5.59      \\ \hline
      &             &            &                        
& \multicolumn{6}{c|}{Middlebury (other)}                                             & \multicolumn{6}{c}{Vimeo90k (interpolation)}                                                \\ \cline{5-16} 
                   \multirow{-1}{*}{Method} & \multirow{-1}{*}{Frame} & \multirow{-1}{*}{Event}& \multirow{-1}{*}{Supervision}  & \multicolumn{3}{c|}{3 skip} & \multicolumn{3}{c|}{1 skip } & \multicolumn{3}{c|}{3 skip} & \multicolumn{3}{c}{1 skip} \\\cline{5-16} 
                         &            &             &                         & PSNR$\uparrow$       & SSIM$\uparrow$      & IE$\downarrow$        & PSNR$\uparrow$       & SSIM$\uparrow$       & IE$\downarrow$        & PSNR$\uparrow$       & SSIM$\uparrow$      & IE$\downarrow$        & PSNR$\uparrow$       & SSIM$\uparrow$       & IE$\downarrow$       \\ \hline
E2VID\cite{rebecq2019high}                    &         \ding{56}        &   \ding{52}&   \ding{52} & 11.26      & 0.427     & 69.73     & 11.82      & 0.403      & 70.15     & -          & -         & -         & 10.08      & 0.395      & 79.89     \\
SepConv\cite{niklaus2017video}                  &         \ding{52}       &    \ding{56}&   \ding{52}            & 25.51      & 0.824     & 6.74      & 30.16      & 0.904      & 3.93      & -          & -         & -         & 33.80      & 0.959      & 3.15      \\
QVI \cite{xu2019quadratic}                     &         \ding{52}       &    \ding{56}&   \ding{52}            & 26.31      & 0.827     & 6.58      & 31.02      & 0.908      & 3.78      & -          & -         & -         & -          & -          & -         \\
DAIN\cite{bao2019depth}                     &         \ding{52}       &    \ding{56}&   \ding{52}            & 26.67      & 0.838     & 6.17      & 30.87      & 0.899      & 4.86      & -          & -         & -         & 34.20      & 0.962      & 3.03      \\
SuperSloMo\cite{jiang2018super}               &        \ding{52}        &     \ding{56}&   \ding{52}           & 26.14      & 0.825     & 6.33      & 29.75      & 0.887      & 4.65      & -          & -         & -         & 32.93      & 0.948      & 3.50      \\
Time Lens\cite{tulyakov_time_2021}                &        \ding{52}        &     \ding{52}&   \ding{52}           & 32.13      & 0.908     & 4.07      & 33.27      & 0.929      & 3.17      & -          & -         & -         & 36.31      & 0.962      & 2.38      \\
\hdashline[4pt/5pt]
UnSuperSloMo\cite{reda2019unsupervised}             &        \ding{52}        &     \ding{56}&   \ding{56}           & 24.86      & 0.789     & 7.40      & 28.27      & 0.873      & 4.98      & -          & -         & -         & 30.38      & 0.930      & 3.77      \\
Ours                     &        \ding{52}        &     \ding{52}&   \ding{56}           & 30.91      & 0.887     & 4.89      & 32.74      & 0.912      & 3.63      & -          & -         & -         & 35.12      & 0.963      & 2.79      \\
\bottomrule
\end{tabular}
   }
   \label{Tab.performance} 
    \label{tab_performance} 
   \end{table*}
\begin{table*}[!t]
  \renewcommand\arraystretch{1.3}
  \centering
  \caption{Quantitative comparison on HQF and HS-ERGB dataset with real-world events.}
  \vspace{-5pt}
  \resizebox{\textwidth}{!}{
  \begin{tabular}{lccccc|cc|cc|cc|cc|cc}
\toprule
                         &              &           &                         & \multicolumn{4}{c|}{HQF}                                                                                     & \multicolumn{4}{c|}{HS-ERGB  (far)}                                                        & \multicolumn{4}{c}{HS-ERGB (close)}\\ \cline{5-16} 
                         &              &           &                         &                                                        \multicolumn{2}{c|}{3 skip}                           & \multicolumn{2}{c|}{1 skip}                           & \multicolumn{2}{c|}{7 skip}                  & \multicolumn{2}{c|}{5 skip}                  & \multicolumn{2}{c|}{7 skip}                  & \multicolumn{2}{c}{5 skip}                  \\ \cline{5-16} 
\multirow{-3}{*}{Method} & \multirow{-3}{*}{Frame} & \multirow{-3}{*}{Event}& \multirow{-3}{*}{Supervision} & PSNR$\uparrow$                     & SSIM$\uparrow$                      & PSNR$\uparrow$                     & SSIM$\uparrow$                      & PSNR$\uparrow$                 & SSIM$\uparrow$                 & PSNR$\uparrow$                 & SSIM$\uparrow$                 & PSNR$\uparrow$                 & SSIM$\uparrow$                 & PSNR$\uparrow$                 & SSIM$\uparrow$                 \\ \hline
E2VID\cite{rebecq2019high}                    & \ding{56}    & \ding{52}& \ding{52}    & 6.70 & 0.315 & 6.70 & 0.315 & 7.01 & 0.372 & 7.05 & 0.374 & 7.68 & 0.427 & 7.73 & 0.432 \\
DAIN \cite{bao2019depth}                    &     \ding{52}         &      \ding{56}& \ding{52}          & 26.10                    & 0.782                     & 29.82                    & 0.875                     & 27.13                & 0.748                & 27.92                & 0.780                & 28.50                & 0.801                & 29.03                & 0.807                \\
SuperSloMo\cite{jiang2018super}               &     \ding{52}         &      \ding{56}& \ding{52}          & 25.54                    & 0.761                     & 28.76                    & 0.861                     & 24.16                & 0.692                & 25.66                & 0.727                & 27.27                & 0.775                & 28.35                & 0.788                \\
RRIN\cite{li2020video}                     &     \ding{52}         &      \ding{56}& \ding{52}          & 26.11                    & 0.778                     & 29.76                    & 0.874                     & 23.73                & 0.703                & 25.26                & 0.738                & 27.46                & 0.800                & 28.69                & 0.813                \\
BMBC\cite{park2020bmbc}                     &     \ding{52}         &      \ding{56}& \ding{52}          & 26.32                    & 0.781                     & 29.96                    & 0.875                     & 24.14                & 0.710                & 25.62                & 0.742                & 27.99                & 0.808                & 29.22                & 0.820                \\
Time Lens\cite{tulyakov_time_2021}                &     \ding{52}         &      \ding{52}& \ding{52}          & 30.57                    & 0.900                     & 32.49                    & 0.927                     & 32.31                & 0.869                & 33.13                & 0.877                & 31.68                & 0.835                & 32.19                & 0.839                \\
\hdashline[4pt/5pt]
UnSuperSloMo\cite{reda2019unsupervised}             &     \ding{52}         &      \ding{56}& \ding{56}          &     23.47                &          0.740            &        26.11                  &             0.852              &    23.80                  &          0.651            &              25.81        &        0.700              &          26.72            &    0.732                  &           28.38           &         0.741             \\
Ours                     &     \ding{52}         &      \ding{52}& \ding{56}          &       28.82              &         0.866             &         31.07                 &    0.931                &     30.07                 &       0.834               &        31.98              &     0.861                 &    29.83                  &            0.816          &            31.21          &        0.818              \\ 
\bottomrule
\end{tabular}
\vspace{-2cm}
  }
  \label{Tab.performance_real} 
  \end{table*}  

   \begin{figure*}[h] 
      \centering 
      \includegraphics[width=1\textwidth]{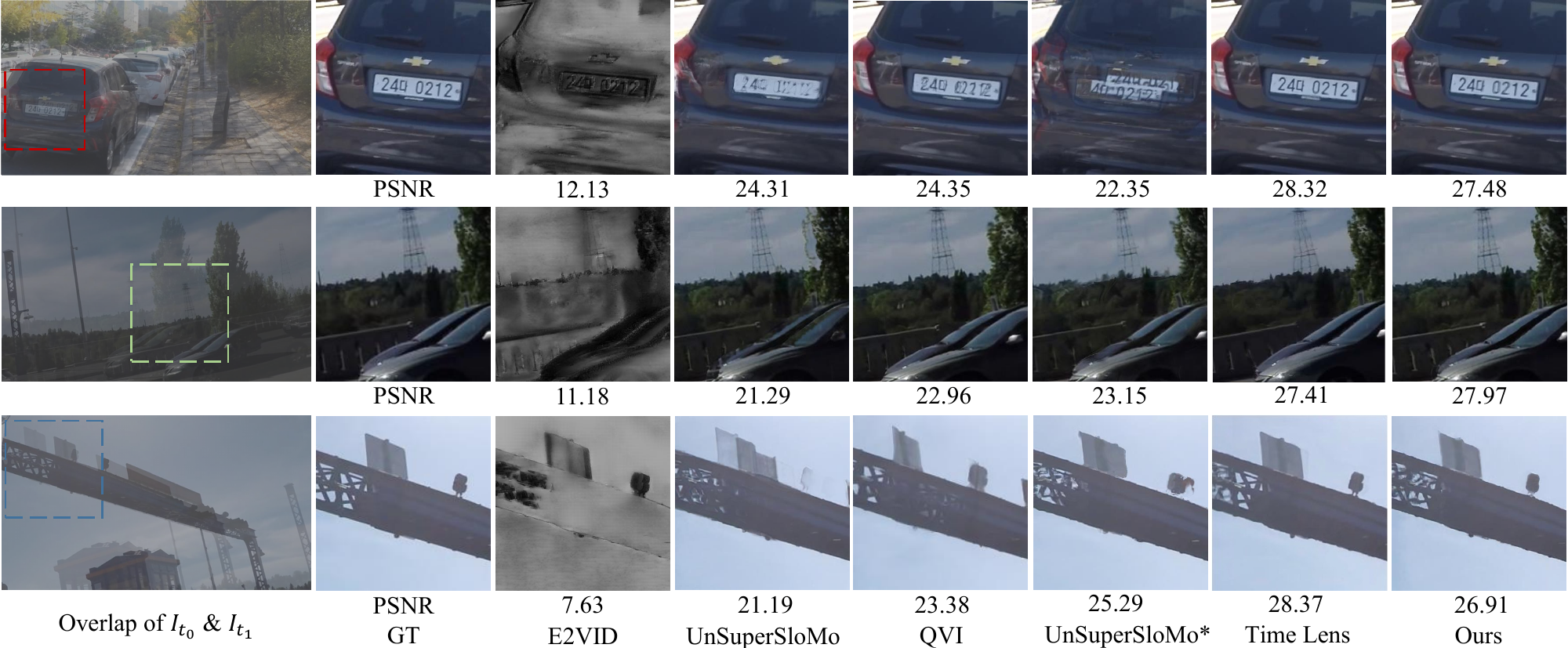}
      \caption{Visual comparison on the Adobe240 and GoPro datasets. In the left, two paired input frames are plotted together; columns in the right are interpolation results from different methods, with a focus on the boxed area.}
      \label{Fig.cherry} 
   \end{figure*}
   \begin{figure*}[h] 
      \centering 
      \includegraphics[width=1\textwidth]{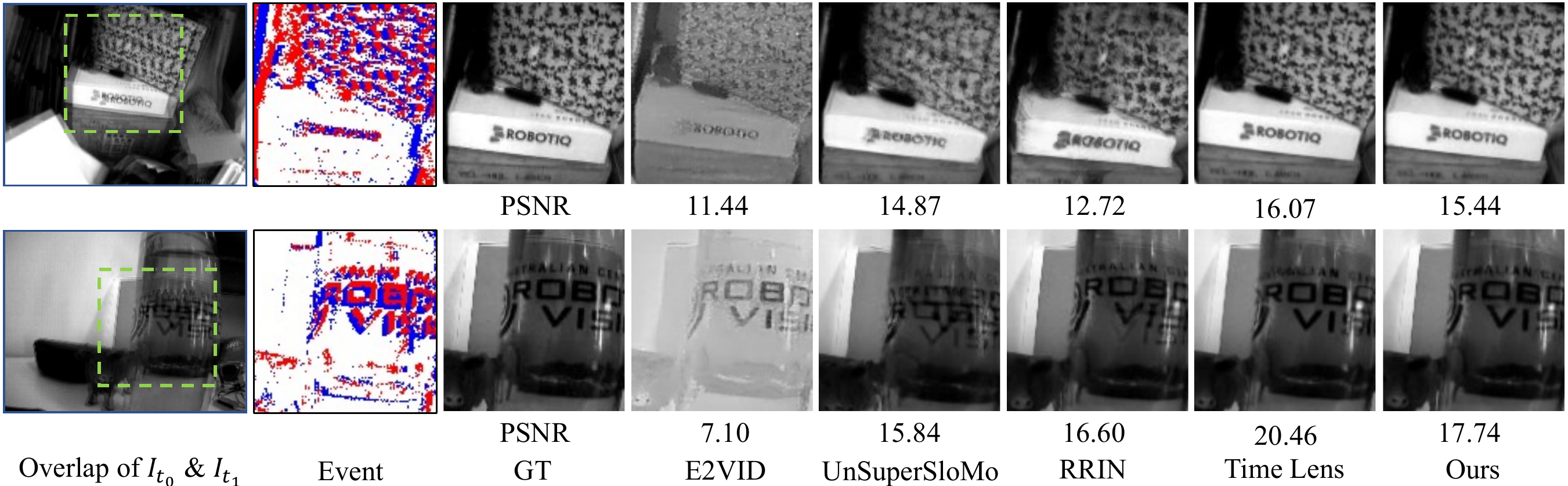}
      \caption{Visual comparison on the HQF dataset. }
      \vspace{-0.5cm}
      \label{Fig.cherry_real_hqf} 
   \end{figure*}

\section{Experiment}

We use Pytorch \cite{paszke2019pytorch} to implement the proposed framework. $256\times 256$ patches are cropped from the corresponding consecutive frames and the batch size is set to 28.
The models are trained using Adam optimizer \cite{kingma2014adam} with with $\beta_1 = 0.9$ and $\beta_2 = 0.999$ on 4 NVIDIA Tesla V100 GPUs. 
The learning rate is initially set to $1e^{-4}$, and then down-scaled by a factor of 0.1 every 200 epochs till 500 epochs. We take peak-signal-to-noise (PSNR), structural-similarity-image-metric (SSIM), and interpolation-error (IE) as a quantitative evaluation metrics for the video interpolation task. Methods with high PSNR and SSIM scores and low IE scores 
are favored in video interpolation. We use arrows to show the favored values, \ie PSNR $\uparrow$, SSIM $\uparrow$, and IE $\downarrow$.

\subsection{Comparison with synthetic events}

\textbf{Training data with synthetic events.} For synthetic data, we collected video sequences from the training set of GoPro\cite{nah2017deep}, Adobe240\cite{su2017deep} and Vimeo90k\cite{xue2019video}. The GoPro dataset contains 22 diverse videos used for training and 11 additional ones for testing, while the Adobe240 dataset has 112 sequences for training and 8 ones for testing. Both video datasets are recorded with GoPro camera at a frame rate of 240 fps with a resolution of $1280\times720$. The Vimeo90k dataset has 51,312 triplets for training, where each triplet contains 3 consecutive video frames with a resolution of $448\times256$ pixels. All the collected video sequences are paired with corresponding synthetic events generated using popular video-to-event simulator ESIM\cite{rebecq2018esim}.

\textbf{Results.} The proposed method is compared with several state-of-the-art interpolation methods, including supervised ones such as Time Lens\cite{Tulyakov21CVPR}, SuperSloMo \cite{jiang2018super}, quadratic video interpolation (QVI) \cite{xu2019quadratic}, depth-aware video frame interpolation (DAIN)~\cite{bao2019depth} and an unsupervised one unsupervised SuperSloMo~\cite{reda2019unsupervised}. Similar to Time Lens, Popular video interpolation benchmark datasets are chosen for evaluating our method, such as GoPro, Adobe240, Vimeo90k (interpolation), and Middlebury (other). For GoPro and Adobe240 datasets, we skip 7 frames of each sequence and form a new sequence as input, and the task reconstructs the skipped ones using the interpolation method. Both average performances over the whole 7 skipped frames and the one of the center skipped one are presented for comparison. As for Vimeo90k (interpolation) and the Other set of Middlebury, 1 or 3 frames are skipped respectively due to the limited sequence length.
For comparison, we simply use the pretrained model on the same dataset provided by the authors. The results are summarised in Tab. \ref{tab_performance}.

Table \ref{tab_performance} shows the quantitative comparison with state-of-the-art methods. 
The method with only event data E2VID performs worst because event data is sparse and only contains relative intensity change, which is not sufficient to reconstruct high-quality intermediate frames.
With the help of event data, methods with both intensity images and event data as input, \ie Time Lens, and the proposed method perform favorably against those methods with only intensity images as input, \ie SuperSloMo, DAIN, SepConv, and QVI. Note that our method, which is trained following an unsupervised fashion, even outperforms most supervised methods. This could be attributed to its powerful capability in accurately modeling nonlinear motion by making full use of almost continuously captured event streams. 

As shown in the first row of Fig.~\ref{Fig.cherry}, when cars are parked on the side of the road and the handheld camera is moving, there is a change with car's position in the image. Only Time Lens and our method, which are able to model motion accurately, can predict a reasonable intermediate frame, especially on license plate. Other methods suffer from the inconsistency of estimation from two input frames, thereby producing artifacts on the numbers. Similar results can also be observed in the other two examples. This demonstrates that event data can indeed help model complex motion with a high-quality intermediate frame interpolation.

\subsection{Comparison on real event}
Experiments above are conducted on synthetic datasets, where event streams are simulated. In this section, we investigate how the proposed method performs on real events in real-world scenarios. Since simulated events are free from background noise and limited read-out bandwidth, there is a gap between simulated and real events. Hence, the performance of methods that are trained on synthetic dataset with the supervision of ground truth may degrade on real data. 

\textbf{Training data with real events.} High Quality Frames (HQF) \cite{stoffregen2020reducing}, and HS-ERGB \cite{tulyakov_time_2021} datasets are chosen for fine-tuning our model trained on synthetic dataset (Vimeo90k). Collected using DAVIS240 event camera, HQF dataset consists of 14 frame sequences with the corresponding event streams in a resolution of $240\times 180$ since both the event channel and the conventional intensity channel are integrated on DAVIS240. HS-ERGB dataset contains sequences from a dual-camera setup, which consists of Prophesee Gen4M 720p event camera and FLIR BlackFly S RGB camera.

\textbf{Results.} 
 Here we evaluate our method on HQF and HS-ERGB datasets\cite{tulyakov_time_2021}, both containing video frame sequences without blur and saturation and the corresponding real-world event streams. Far-away and close planar scenes are divided in HS-ERGB dataset for testing since different rectifications are applied. The quantitative results are summarised in Tab. \ref{Tab.performance_real}. Similar results as the synthetic experiment can be conducted by the quantitative comparison.
As shown in Fig.~\ref{Fig.cherry_real_hqf}
, methods without event input produce many artifacts, especially on the edges. The words in the frame could be found suffering when reconstructing frame sequences using frame-only interpolation methods while avoiding artifacts with the help of event input (Time Lens and ours).  

\subsection{Ablation study}


   In this section, we investigate the contribution of each component in the proposed approach on the GoPro dataset and report experimental results in Table \ref{Tab.ablation}. Three variants are: (1) ``linear motion model'': the optical flows between two input frames are first estimated with both intensity images and event data, and then the flow from two input frames to the target intermediate frame is computed as a proportion of the previously estimated flow. It aims to validate the importance of accurate motion modeling and the effectiveness of the proposed method in modeling complex nonlinear motion. (2) ``shared flow est module'': to investigate the influence caused by the difference between actual event stream and approximately reversed one to the whole framework, we experiment with only one optical flow estimator which is used to deal with both of them. (3) ``UnSuperSloMo*'': to demonstrate the effectiveness
   of the proposed method, we also compare it with the enhanced
   version of unsupervised SuperSloMo,
   where the four-channel event representation is taken as additional input and concatenated with input images before being fed to the model. As shown in Table~\ref{Tab.ablation}, all these components contribute to the performance of video interpolation. 

\begin{table}[h]
   \renewcommand\arraystretch{1.3}
   \centering
   \caption[]{Ablation results on GoPro (7 skip).}
   \vspace{-0.2cm}
   \resizebox{0.85\columnwidth}{!}{
      \begin{tabular}{ccccccc}
         \toprule
Method
                                 & & PSNR$\uparrow$   & & SSIM$\uparrow$   & & IE$\downarrow$    \\ \hline
         linear motion       & &  32.05      & &   0.939     & &6.40  \\
         shared flow est module     & &  33.78      & &    0.958    & &    5.16   \\
         UnSuperSloMo*       & &  32.16      & &   0.942     & &6.31\\

         Ours                    & & 34.02  & & 0.960  & & 5.02  \\
         \bottomrule
         \vspace{-1cm}
   \end{tabular}
   }
   \label{Tab.ablation} 
   \end{table}

\subsection{Benefit of unsupervised VFI with events}

An important advantage of our proposed method is its unsupervised nature. Previous methods generate high quality ground truth frames for training VFI model in a supervised manner. They require either carefully controlled slow shooting like HQF~\cite{stoffregen2020reducing}, or complex registration among high-speed cameras and event cameras~\cite{tulyakov_time_2021}, which can not be easily obtained with low cost. In contrast, an unsupervised approach can benefit from arbitrary videos and paired events, which could be easy obtained at scale via DAVIS sensor. And the large scale unsupervised training data could further improve the VFI performance.

To demonstrate the benefit of unsupervised training, we select the large scale DDD-17 dataset~\cite{binas2017ddd17} for unsupervised finetuning, which containing over 12h record of a 346x260 pixel DAVIS sensor. DAVIS sensor produces low-speed frames together with paired events, which cannot be used by previous methods due to the lack of high-speed intermediate frames. Since our TimeReplayer algorithm relies on unsupervised cycle-consistent training, DDD-17 dataset can be used for unsupervised finetuning. The results are summarized in Tab. \ref{Tab.performance_real_supp}.

\begin{table}[h]
  \centering
  \caption{Quantitative comparison on HQF dataset with real events.}
  \vspace{-5pt}
  \resizebox{\columnwidth}{!}{
  \begin{tabular}{lll|ll}
\toprule       &                                                        \multicolumn{2}{c}{3 skip}                           & \multicolumn{2}{c}{1 skip}                                             \\ \cline{2-5} 
\multirow{-2}{*}{Method}  & PSNR$\uparrow$                     & SSIM$\uparrow$                      & PSNR$\uparrow$                     & SSIM$\uparrow$                      \\ \hline

Time Lens-syn        & 28.98                    & 0.873                     & 30.57                    & 0.903\\
Time Lens-real        & 30.57(\color{red}$+1.59$)                    & 0.900(\color{red}$+0.027$)                     & 32.49(\color{red}$+1.92$)                    & 0.927(\color{red}$+0.024$)\\

Ours          &       28.82              &         0.866             &         31.07                 &    0.931                \\
Ours+DDD17        & 31.54(\color{red}$+2.72$)                   & 0.920(\color{red}$+0.054$)      &  33.93(\color{red}$+2.86$)    & 0.934(\color{red}$+0.003$) \\
\bottomrule
\end{tabular}
  }
  \vspace{-10pt}
  \label{Tab.performance_real_supp} 
  \end{table}  

To get better performance, Time Lens~\cite{tulyakov_time_2021} needs to be fine-tuned with high-cost real event data. In contrast, our method can be fine-tuned using \emph{low-cost unsupervised event data}, which surpasses Time Lens thanks to the abundance of data in DDD17.

\section{Conclusion}
In this paper we introduce event camera into the video interpolation framework to help reconstruct intermediate frames at the presence of complex motion. We propose the TimeReplayer algorithm, a cycle-consistency based unsupervised training method which allows a video interpolation model to be trained in an unsupervised way with only two frames and events. It reduces the requirement of large amounts of paired high-frame-rate videos and unlocks the potential of event data in video frame interpolation. Natural data collected by DAVIS camera at a low cost can be used to improve the performance of TimeReplayer. Extensive experiments on both synthetic datasets and real data demonstrate the effectiveness of the proposed method in addressing complex nonlinear motion and reconstructing high-quality intermediate frames. 

\section{Acknowledgements}
This work was supported by the National Key Research and Development Program of China (Grant No. 2021ZD0200300), and the Natural Science Foundation of China, No. 62106036, 61725202, U1903215, and the Fundamental Research Funds for the Central University of China, DUT No. 82232026.
\newpage
{\small
\bibliographystyle{ieee_fullname}
\bibliography{egbib}
}

\end{document}

%% file: pre_and_method_v2.tex
\section{Method}
\label{method}

In this section, we first describe the representation of event data used in this framework and then explain how event stream contributes to the video interpolation process in case of complex nonlinear motion. Finally, we describe the proposed unsupervised training framework for video interpolation with event data.

\subsection{Event Representation.} 

Event camera asynchronously captures log intensity change $\Delta log(I_{x,y,t})$ for each pixel and an event $\mathcal{E}_{x,y,t,p}$ is triggered once the change at position $(x, y)$ and timestamp $t$ reaches a certain threshold $\tau$. Therefore, it is able to "continuously" capture intensity change with extreme low latency (in the order of $\mu s$). The polarity $p$ of an event means the sign of $\Delta log(I_{x,y,t})$,
with $1$ and $-1$ representing the positive and negative events respectively. Event stream is a sequence of events within the time period $[t_{0},t_{0}+T]$ containing $N$ events: $e_{t_{0}\rightarrow (t_{0}+T)} = \{\mathcal{E}_{x_{i},y_{i},t_{i},p_{i}}\}_{i = 0}^{N-1}$.

The raw form of the spatial-temporal event stream is a sequence of '$1$'s and '$-1$'s at different time stamps, which is not friendly to standard neural network models such as CNNs, hence not suitable for direct processing by those networks. Raw event stream needs to be first transformed into an image-like input. 
In this work event stream is represented as a 4-channel image with the first two channels encoding counts of positive and negative events at each pixel, 
while the other two channels encoding corresponding to timestamps of the latest triggered positive/negative events at each pixel~\cite{7532633}.

 \begin{figure*}[t] 
   \centering 
   \includegraphics[width=0.9\textwidth]{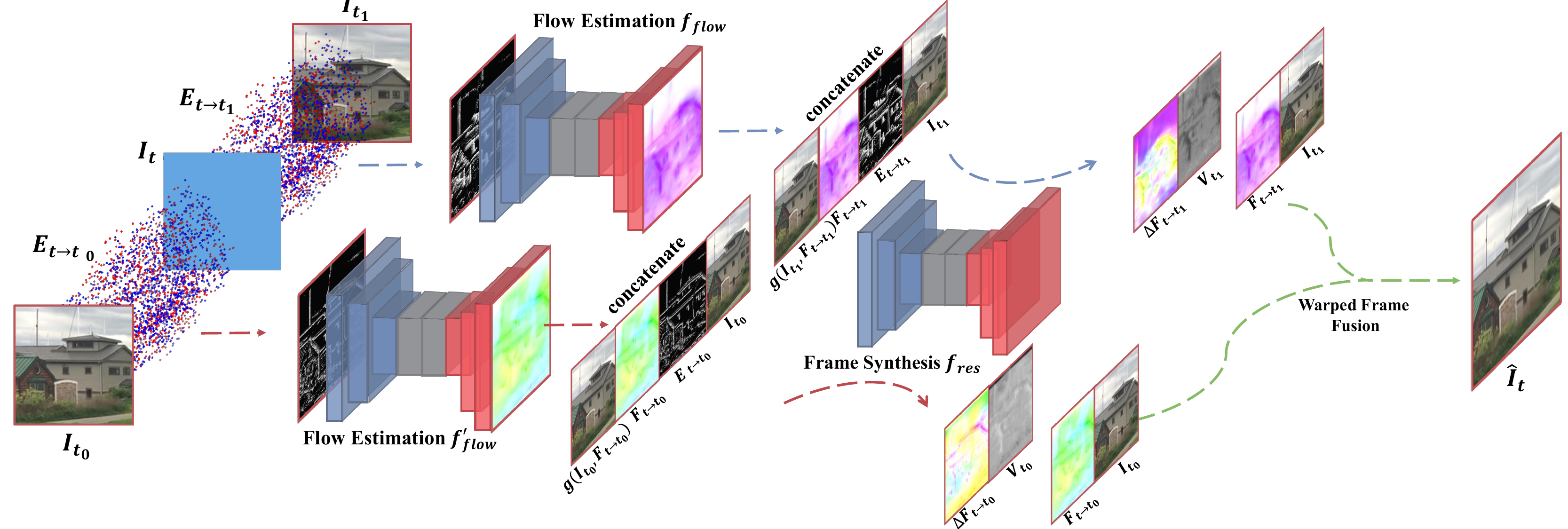}
   \caption{The proposed TimeReplayer model architecture with event stream. 
   } 
   \label{Fig.networks} 
\end{figure*}

As a directional process quantity, the event stream $e_{t_{0}\rightarrow t_{0}+T}$ could be approximately reversed by simultaneously exchanging the positive/negative polarity of events and changing the order of timestamps, whose use would be detailed in video interpolation model. 
The approximation is referred to as $e^\prime_{(t_{0}+T)\rightarrow t_{0}} = \{\mathcal{E}_{x_{i},y_{i},t_{i},-p_{i}}\}_{i = N-1}^{0}$.

\begin{figure}[ht] 
   \centering 
   \includegraphics[width=0.45\textwidth]{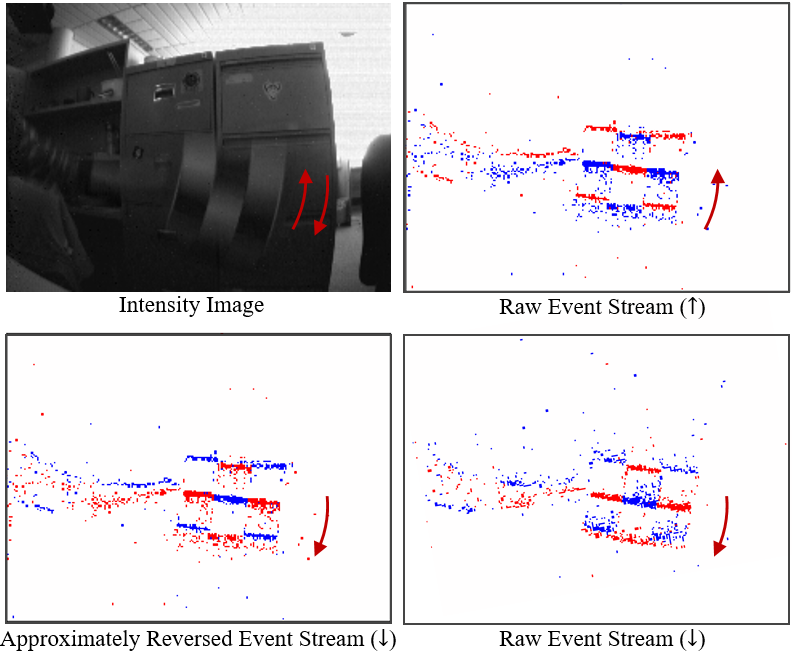}
   \caption{Visualization of reverse events. We can see that although approximately reversed events for downward motion look similar to the actual events for downward motion, there is still much difference between them, especially on non-edge regions.
   }
   \label{Fig.reverse} 
\end{figure}

In fact, due to the characteristics of the hardware circuit of the event camera, the approximately reversed event stream 
$e^\prime_{(t_{0}+T)\rightarrow t_{0}}$
 is not exactly the same as the actual backward streaming events. 
 An example is shown in Fig. \ref{Fig.reverse}. Therefore we propose to process the original forward streaming event streams and the approximately reversed ones  separately. We use $E_{t_{0}\rightarrow t_{1}}$ to refer to the event stream from $t_{0}$ to $t_{1}$ as $E_{t_{0}\rightarrow t_{1}} = \left\{\begin{matrix}
   e_{t_{0}\rightarrow t_{1}}, \; if \; t_{0}\leq t_{1}\\
   e'_{t_{0}\rightarrow t_{1}},\;  if \; t_{0} > t_{1}
   \end{matrix}\right.$

\subsection{Video Interpolation Model}
The proposed model is based on SuperSloMo~\cite{jiang2018super} with three modules, which are described in the following.

\textbf{Flow estimation.} Since event stream inherently carries continuous motion information with itself, optical flow can be directly computed by feeding event representation to a convolutional neural network. However, due to the difference between approximately reversed event stream and the actual one, we take two flow estimation CNN modules $f_{flow}$ and $f^\prime_{flow}$ , which have the same network architecture but do not share parameters, to deal with forward streaming events $E_{t\rightarrow t_1}$  and backward streaming ones $E_{t\rightarrow t_0}$ respectively. The correspondingly computed optical flows for $E_{t->t_1}$ and $E_{t->t_0}$ are respectively denoted as $F_{t\rightarrow t_1}$ and $F_{t\rightarrow t_0}$. 
Compared to most previous methods, which assume that objects move along a straight line at a constant speed and compute the optical flow between an input frame and the targeted intermediate frame as a proportion of the optical flow between two input frames, this method can make a more accurate estimation of optical flow between the interpolated frame and input frames in case of nonlinear motion.

\textbf{Flow refinement.} However, event-based optical flow estimation methods tend to produce sparse output concentrating on edges of moving objects. Thus, another CNN $f_{flow\_res}$ is used to refine the initially estimated sparse optical flows $F_{t\rightarrow t_0}$ and $F_{t\rightarrow t_1}$ by computing a residual flow $\Delta F$. The final estimation is obtained by taking the sum of the initial estimation and a residual.
\begin{equation}
   \begin{aligned}
      (\Delta F_{t\rightarrow t_0},V_{t_0}) = f_{res}(g(I_{t_0},F_{t\rightarrow t_0}),I_{t_0},F_{t\rightarrow t_0}, E_{t\rightarrow t_0})\\
      (\Delta F_{t\rightarrow t_1},V_{t_1}) = f_{res}(g(I_{t_1},F_{t\rightarrow t_1}),I_{t_1},F_{t\rightarrow t_1}, E_{t\rightarrow t_1}),
   \end{aligned}
   \label{eq:deltaF}
\end{equation}
where $g(I_{t_0},F_{t\rightarrow t_0})$ and $g(I_{t_1},F_{t\rightarrow t_1})$ are warped input frames computed using initially estimated optical flows $F_{t\rightarrow t_0}$ and $F_{t\rightarrow t_1}$, $\Delta F_{t\rightarrow t_0}$ and $\Delta F_{t\rightarrow t_1}$ denote the residual used to refine the optical flows, and $V_{t_0}$ and $V_{t_1}$ represent visibility maps which help alleviate artifacts and occlusion in the later blending process. 
Two sets of feature maps are separately concatenated and passed through a shared optical flow refinement model $f_{flow\_res}$.

\textbf{Frame synthesis.}
Finally the target frame $\hat{I}_{t}$ can be synthesized by blending the warped input frames using refined optical flows. The blending process is taken as a weighted average of two warped frames with the product of time interval and visibility map as weights, as described in Eq.~\ref{eq:blend}.
\begin{equation}
   \begin{aligned}
      \hat{I}_{t} = &\frac{1}{Z}\odot((t_1-t)V_{t_0}\odot g(I_{t_0},F_{t\rightarrow t_0}+\Delta F_{t\rightarrow t_0})+\\&(t-t_0)V_{t_1}\odot g(I_{t_1},F_{t\rightarrow t_1}+\Delta F_{t\rightarrow t_1})).
   \end{aligned}
   \label{eq:blend}
\end{equation}
where $Z$ is a normalization term computed as the sum of $(t_1-t)V_{t_0}$ and $(t-t_0)V_{t_1}$.

Therefore, as long as two input frames at two time stamps $t_0$ and $t_1$, and the event streams between these two time stamps and the targeted one $t$ are given, we could synthesize the desired frame at that time stamp with the proposed whole video interpolation model $f$.

\subsection{Temporal Cycle Consistency}
\label{sec:cycle_training}
In this work we focus on training a video interpolation model in an unsupervised fashion\cite{reda2019unsupervised}. Inspired by cycle consistency, which is first proposed in CycleGAN \cite{zhu2017unpaired} and UNIT \cite{liu2017unsupervised} and then becomes a popular constraint in self-supervised learning, we present an unsupervised training framework for video interpolation with event data. 
As explained in previous section, with the help of event data, the model is able to synthesize an intermediate frame between two input frames at different time stamps; similarly, it can also synthesize an input frame given the predicted intermediate frame and another input frame. In this way, we can impose supervision on the reconstruction of input frames to train the video interpolation model.  
Specifically, as for an interval between time $t_0$ and $t_1$, given a pair of input frames $I_{t_0}$ and $I_{t_1}$, and event stream $E_{t_0\rightarrow t_1}$, then we can use the video interpolation model $f$ to produce an intermediate frame $\hat{I}_{t}$ in-between as $\hat{I}_{t} = f(I_{t_0}, I_{t_1}, E_{t\rightarrow t_0}, E_{t\rightarrow t_1})$,
where $f$ denotes the whole video interpolation model, $t\in (t_0, t_1)$ denotes the target interpolation time, and $E_{t\rightarrow t_0}$ and $E_{t\rightarrow t_1}$ respectively represent approximately reversed streaming events from time $t$ to time $t_0$ and the actual forward streaming ones from $t$ to time $t_1$.

Similarly, with $t_1$ as target time, given a pair of input frames $I_{t_0}$ and $\hat{I}_{t}$, and the corresponding event streams from time $t_1$ to $t_0$ and from time $t_1$ to $t$, the same video interpolation model $f$ could be applied to reconstruct frame $I_{t_1}$ as $\hat{I}_{t_1} = f(\hat{I}_{t}, I_{t_0}, E_{t_1\rightarrow t}, E_{t_1\rightarrow t_0}) $,
where $E_{t_1\rightarrow t_0}$ and $E_{t_1\rightarrow t}$ are approximately reversed event streams.

With $t_0$ as target time, given a pair of input frames $I_{t_1}$ and $\hat{I}_{t}$, and the corresponding event streams from time $t_0$ to $t_1$ and from time $t_0$ to $t$, the same video interpolation model $f$ could also be applied to reconstruct frame $I_{t_0}$ as $\hat{I}_{t_0} = f(\hat{I}_{t}, I_{t_1}, E_{t_0\rightarrow t}, E_{t_0\rightarrow t_1})$, where $E_{t_1\rightarrow t_0}$ and $E_{t_1\rightarrow t}$ are the corresponding actual event streams.

\begin{figure}[t] 
   \centering 
   \includegraphics[width=0.48\textwidth]{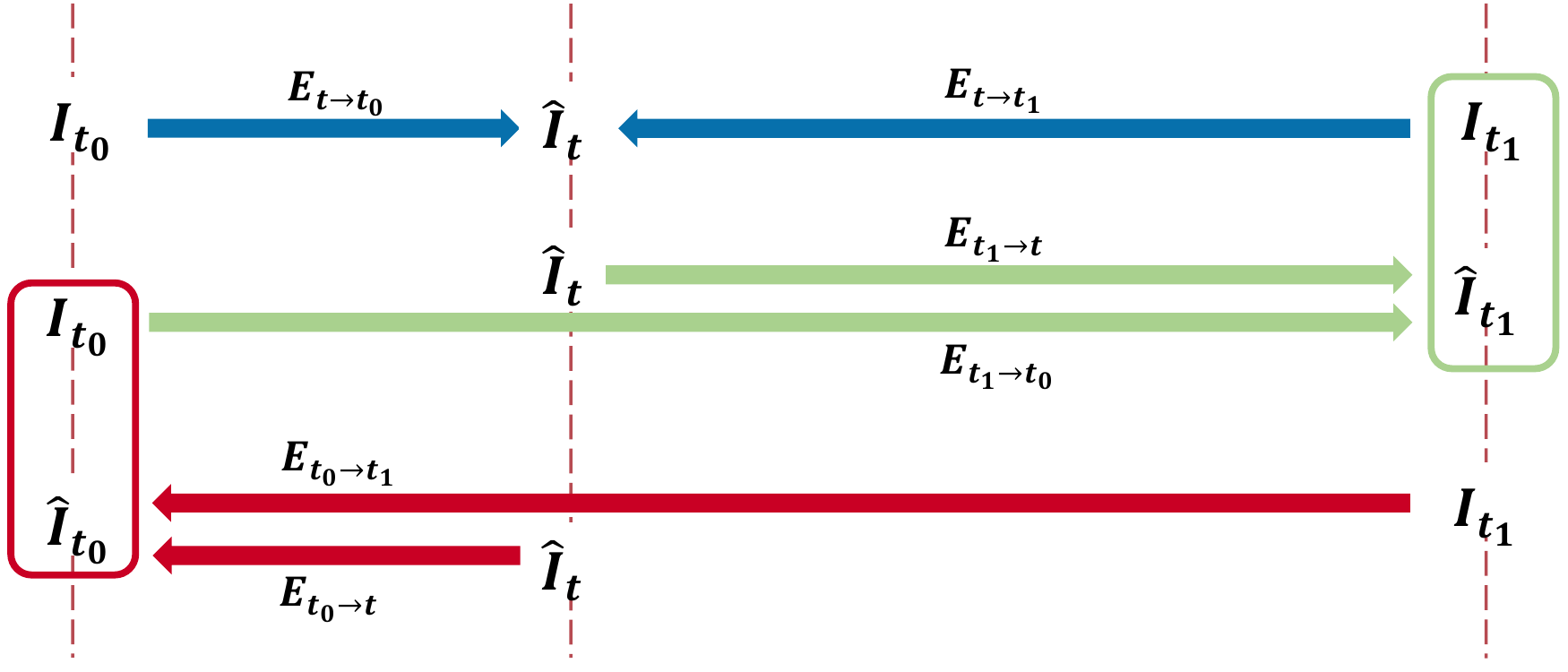}
   \caption{Temporal cycle consistency.  $I_{t_0}$ and $I_{t_1}$ are the original input frames, and $E_{t_0\rightarrow t}$ and $E_{t\rightarrow t_1}$ are input event streams. After obtaining an estimated intermediate frame $\hat{I}_{t}$, we could further apply the same video interpolation model to reconstruct $I_{t_1}$ and $I_{t_0}$, which can be used for imposing cycle-consistent learning.}
   \label{fig:architecture} 
\end{figure}

The interpolation procedure is visualized in Fig.~\ref{fig:architecture}.

In this way, the estimated intermediate frames $\hat{I}_{t_0}$ and $\hat{I}_{t_1}$ could be respectively paired with the original input frames $I_{t_0}$ and $I_{t_1}$. Therefore we can train the interpolation model $f$ by minimizing the reconstruction error between them, \ie cycle consistency loss as $\mathcal{L}_{cc} = ||\hat{I}_{t_0}-I_{t_0}||_{1}+||\hat{I}_{t_1}-I_{t_1}||_{1}$.

The final loss function integrates the above terms $\mathcal{L}_{total} = \lambda_c \mathcal{L}_{cc} + \lambda_w \mathcal{L}_{warp} + \lambda_s \mathcal{L}_{smooth} + \lambda_p \mathcal{L}_{percep},$
where warping loss $\mathcal{L}_{warp}$, smootheness loss $\mathcal{L}_{smooth}$ and perceptual loss $\mathcal{L}_{percep}$ are defined in a similar way as in SuperSloMo~\cite{jiang2018super} but for both predictions at $t_0$ and $t_1$.